\documentclass[runningheads]{llncs}
\usepackage{graphicx}
\usepackage{wrapfig}
\usepackage{amsmath,amssymb} %
\usepackage{color}

\begin{document}
\title{Labeling Panoramas with\\Spherical Hourglass Networks} 

\titlerunning{Labeling Panoramas with Spherical Hourglass Networks}
\author{\small Carlos Esteves\textsuperscript{1,2}, Kostas Daniilidis\textsuperscript{1}, Ameesh Makadia\textsuperscript{2}}
\institute{\small \textsuperscript{1}GRASP Laboratory, University of Pennsylvania \quad \textsuperscript{2}Google
{\small \{machc,kostas\}@seas.upenn.edu} \quad makadia@google.com}
\authorrunning{C. Esteves, K. Daniilidis and A. Makadia}
\maketitle              %
\begin{abstract}
With the recent proliferation of consumer-grade $360^\circ$ cameras, it is worth revisiting visual perception challenges with spherical cameras given the potential benefit of their global field of view. To this end we introduce a spherical convolutional hourglass network (SCHN) for the dense labeling on the sphere. The SCHN is invariant to camera orientation (lifting the usual requirement for ``upright'' panoramic images), and its design is scalable for larger practical datasets. Initial experiments show promising results on a spherical semantic segmentation task.
\end{abstract}
\section{Introduction}
Panoramic sensors have long been favored for tasks that benefit from $360^\circ$ field of views. For example, omnidirectional sensing for robotic navigation was explored as early as~\cite{yagi90iros}, and panoramic images have provided the building blocks for early VR environments~\cite{chen95siggraph}. While the hardware profile of the early imaging devices limited their broad adoption (e.g. mirror-lens catadioptric sensors~\cite{nayar97cvpr}),
recent hardware and algorithmic advances have created a proliferation of consumer-grade $360^\circ$-cameras (e.g. the Garmin\textsuperscript{TM} VIRB 360 and Samsung Gear\textsuperscript{TM} 360). With the resulting surge in panoramic image datasets, it is natural to investigate machine learning solutions for visual perception tasks on the sphere.

In this work we explore the task of supervised dense labeling for spherical images. CNNs have provided breakthroughs for the analogous 2D image pixel-labeling tasks~\cite{badrinarayanan15cvpr,ronneberger15miccai,long15cvpr}.
 The difficulty in applying these models directly to spherical images (e.g. equirectangular projections of the sphere) is that the usual 2D grid convolution will not reflect the underlying spherical topology (i.e. the convolutions would be artificially susceptible to the distortions that are byproduct of any sphere-to-plane projection).

In contrast, we consider true spherical convolutions executed in the spectral domain~\cite{arfken1966mathematical}.
Specifically, we introduce the Spherical Convolutional Hourglass Network (SCHN), which leverages spherical residual bottleneck blocks arranged in an encoder-decoder style hourglass architecture~\cite{newell2016stacked} to produce dense labels in an $SO(3)$-equivariant fashion. The contributions of this work are:
\begin{itemize}
\item a model for pixel-labeling equivariant to 3D rotations, thus relaxing constraints on camera orientation without data augmentation.
\item a scalable design with an hourglass structure along with localized spherical filters controlled by very few spherical Fourier coefficient parameters.
\end{itemize}

\section{Related Work}
\subsubsection{Panoramic scene understanding}\hfill\\
Recent efforts on scene understanding include PanoContext~\cite{zhang14eccv} (3D room layouts from panoramas) and~\cite{deng17iccar} (applying R-CNN~\cite{girshick15iccv} directly to the spherical panoramas).
Im2Pano3D~\cite{song2016im2pano3d} proposes CNNs for semantic extrapolation from a partial view.
In the examples above, in contrast to our proposal, there is a limiting assumption about the camera orientation which implicitly requires all images to be in an ``upright'' pose.  Circumventing this in the context of deep networks may not be practical, as data augmentation would require sampling from the entire group of 3D rotations as well as networks with much larger capacity.
\subsubsection{Deep learning on the sphere}\hfill\\
Deep learning for spherical images requires models that are faithful to the underlying spherical topology. The spherical CNN~\cite{s.2018spherical} utilizes spherical correlation to map inputs to features on the rotation group $SO(3)$ which are subsequently processed $SO(3)$-convolutional layers. An alternative is a network composed only of spherical convolutional layers~\cite{esteves_sphcnn}. In~\cite{kondor_cbnet} the Clebsch-Gordan decomposition is used to develop a spectral nonlinear activation function to allow fully spectral networks. In~\cite{su17nips} a custom 2D convolution on the equirectangular image correctly accounts for the projection distortion and is efficiently executed, however it is not rotation equivariant. Our proposed architecture is composed of the convolutional layers from~\cite{esteves_sphcnn} as this is the most efficient of the $SO(3)$-equivariant models while still producing state-of-the-art results on 3D shape classification.

\section{Spherical hourglass networks}
\begin{figure}[t]
\includegraphics[width=\textwidth]{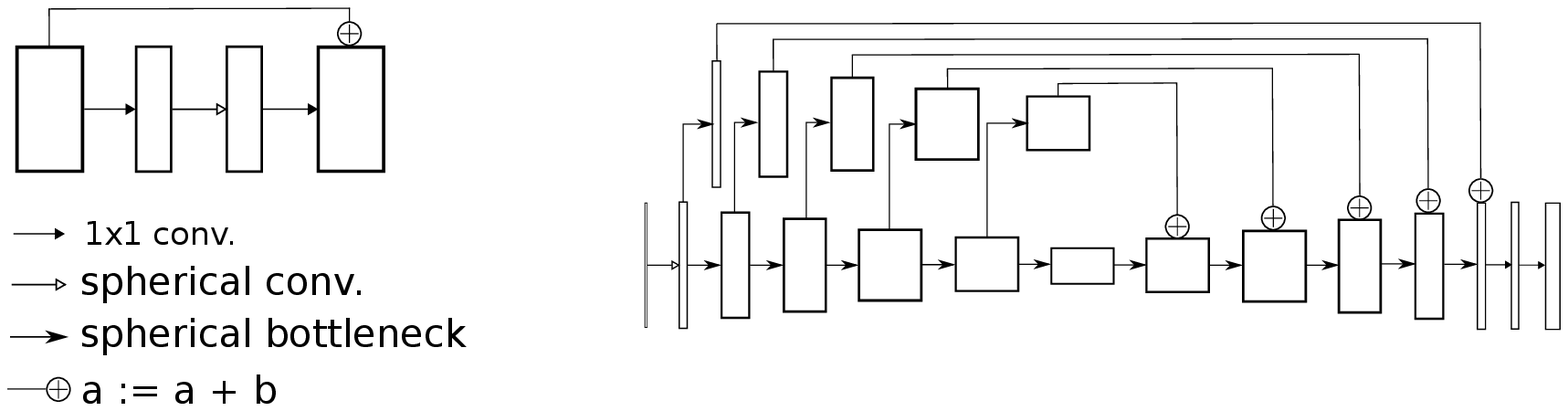}
\caption{\small Network architecture. 
Blocks represent feature maps and arrows, operations. 
The height of a block represent the spatial resolution and the width, the number of channels.
Left: spherical residual bottleneck block.
Right: spherical hourglass network.}
\label{sphhg}
\end{figure}

Due to their high computational cost, equivariant Spherical CNNs~\cite{s.2018spherical,esteves_sphcnn} have only been applied to relatively simple problems like shape classification where shallow networks are sufficient.
For dense labeling, deeper and more sophisticated architectures are needed. %
The outline of our architecture resembles an hourglass, with a series of downsampling blocks followed by upsampling blocks, enabling generation of high resolution outputs.
One key observation of \cite{He_2016_CVPR} is that a residual block with two $3\times 3$ convolutional layers can be replaced by a bottleneck block with $1\times 1$, $3\times 3$, and $1\times 1$ layers, saving compute and increasing performance. 
Since $1\times 1$ convolutions are pointwise operations, hence $SO(3)$-equivariant, we can apply the same idea to spherical convolutional layers, which yield the spherical residual bottleneck blocks (Fig.~\ref{sphhg}).

Both dilated~\cite{yu2015multi} and deformable~\cite{dai2017deformable} convolutions have proven useful for semantic segmentation and the spherical filters we use share some of their properties. As explained in~\cite{esteves_sphcnn}, the number of anchor points in the spectrum loosely determines their receptive field, as in a dilated convolution. While the number of anchor points is fixed and small, ensuring some smoothness in the spectrum (and thus spatial localization), the weights learned at these anchors can provide some measure of deformation to the support.

\section{Experiments}
\begin{wraptable}{r}{0.4\textwidth}\vspace{-30pt}
\caption{\small Mean IoU.
\textit{c} is canonical orientation, \textit{3d} is arbitrary.
}\label{tab1}
\centering
{\scriptsize

\begin{tabular}{|lccc|}
\hline
& \multicolumn{3}{c|}{train/test orientation} \\
 & c/c & 3d/3d & c/3d\\
\hline
SCHN (ours)  &  0.5683 & \textbf{0.5582} & \textbf{0.5024} \\
2DHG &  \textbf{0.6393} & 0.5292 & 0.2237 \\
SCHN/global &  0.5343 & 0.5376 & 0.4758\\
SCHN/large &  0.5983 & 0.5873 & -\\
Im2Pano3D \cite{song2016im2pano3d} &  0.330\footnotemark  &  - & -\\
\hline
\end{tabular}
}
\vspace{-20pt}
\end{wraptable}
\footnotetext{Results for a much harder extrapolation problem. Included here for reference only.}
For segmentation experiments we use all the labeled panoramas from~\cite{song2016im2pano3d} (rendered from ~\cite{song2016ssc}).
We map each sky-box image onto the sphere, with a train-test split of 75k-10k.

The input size is $256\times 256$, we use between 32 and 256 channels per layer and 16 anchor points for filter localization.
We create a 2D baseline (2DHG) which has the exact same architecture of SCHN, with the spherical convolutions replaced by $3\times 3$ 2D convolutional layers. 
Table \ref{tab1} shows the results after training for 10 epochs. SCHN outperforms the baseline under arbitrary orientations, localized filters outperform global, and using larger models can improve the SCHN performance. 

\begin{figure}
\includegraphics[width=\textwidth]{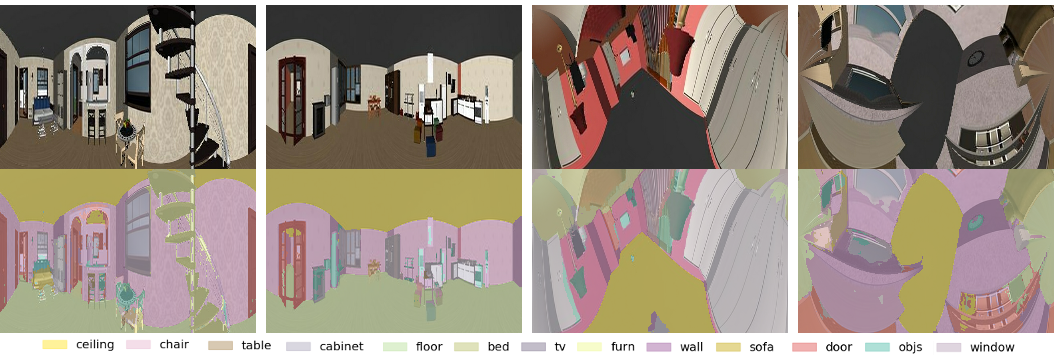}
\caption{\small Top: input spherical panoramas. 
Bottom: segmentation masks produced by our network. 
The leftmost two are in canonical orientation, the others are arbitrary.} 
\label{sample_outputs}
\end{figure}

\section{Conclusion}
SCHNs are the first rotation-equivariant models scalable to learning pixel labeling in spherical images. Promising initial results for semantic segmentation merit further exploration on larger real labeling challenges.
\bibliographystyle{splncs}
{\small
\bibliography{refs.bib}
}
\end{document}